# Streamlined Global and Local Features Combinator (SGLC) for High Resolution Image Dehazing

Bilel Benjdira[1]   Anas M. Ali[1]   Anis Koubaa[1]

[1]RIOTU Lab, Prince Sultan University, Saudi Arabia

{bbenjdira, aaboessa, akoubaa}@psu.edu.sa

**Abstract**

*Image Dehazing aims to remove atmospheric fog or haze from an image. Although the Dehazing models have evolved a lot in recent years, few have precisely tackled the problem of High-Resolution hazy images. For this kind of image, the model needs to work on a downscaled version of the image or on cropped patches from it. In both cases, the accuracy will drop. This is primarily due to the inherent failure to combine global and local features when the image size increases. The Dehazing model requires global features to understand the general scene peculiarities and the local features to work better with fine and pixel details. In this study, we propose the Streamlined Global and Local Features Combinator (SGLC) to solve these issues and to optimize the application of any Dehazing model to High-Resolution images. The SGLC contains two successive blocks. The first is the Global Features Generator (GFG) which generates the first version of the Dehazed image containing strong global features. The second block is the Local Features Enhancer (LFE) which improves the local feature details inside the previously generated image. When tested on the Uformer architecture for Dehazing, SGLC increased the PSNR metric by a significant margin. Any other model can be incorporated inside the SGLC process to improve its efficiency on High-Resolution input data.*

## 1. Introduction

The Image Dehazing problem is attracting an increasing research interest due to the necessity to understand images captured in hazy scenes. Many computer vision tasks such as classification, object detection, tracking, and semantic segmentation will fail in such scenarios. The complexity of the haze particles and their non-homogeneity represent the main challenges faced. When light scatters through the haze, both blur and degradation are applied non-uniformly to many parts of the scene. This pushes to design new specific models that can recover the authentic visual form of the image and make it useful for other tasks. Non-uniform haze is the case in real scenarios, but for simplicity, many research papers began by working on simple homogeneous haze [1]–[4]. In reality, the physical model representing the effect of haze particles on the absorption, scattering, and attenuation of the light is represented by this traditional physical model[5]–[7]:

$$I_h(x) = t(x)I_o(x) + (1 - t(x))A(x) \qquad (1)$$

Where $I_h$ is the hazy image captured by the sensor, $I_o$ is the original image with the haze application, $t(x)$ is the medium transmission map incorporating the visual effect of the haze particles, $x$ is the pixel coordinates, and $A(x)$ is the global atmospheric light. The transmission map is expressed as follows:

$$t(x) = e^{-\beta d(x)} \qquad (2)$$

Where $\beta$ is the atmospheric scattering parameter and $d(x)$ is the scene depth which variate following the $x$ coordinates. The preliminary works in Image Dehazing tried to approximate the atmospheric scattering by the intermediate of prior-based methods [8]–[10]. Hand-crafted priors are used to estimate $A(x)$ and $t(x)$, such as non-local prior or dark-channel prior. However, the atmospheric haze distribution is more complicated and not only correlated to the image depth. It depends on other factors that are more difficult to formulate. Hence, the emergence of deep learning [11] has opened the door to new possibilities that seemed previously impossible. Building Image Dehazing models based on deep learning approaches has significantly improved accuracy [12]–[18]. Although these methods achieve state-of-the-art results on the Dehazing task, the computational cost is still huge. Most of them are well-trained for small to medium image sizes and fail to be applied to High-Resolution images[19]. Although some models can be run on low computational resources, most well-performing models are computationally expensive and cannot manage large image sizes[20]. Increasing the accuracy using a lightweight model is becoming more and more challenging[21]–[23] due to the increasing complexity of modern architectures as well as the incorporation of Vision Transformers[24], [25] on them. The primary inherent source of failure is the challenging mission to combine information obtained from global features with the information obtained from local features[19]. This problem is easier to solve in small-sized images as the model works with the whole image as a fixed input. However, when the size increases, the model needs to work on a downscaled version of the image or on cropped patches from it. In these two methods, the accuracy will drop as the global and local features are unbalanced. This study aims to keep the same efficiency of End-To-End models when handling High-Resolution images. We propose the SGLC (*Streamlined Global and Local Features Combinator*) to make any Dehazing model works efficiently on any input image size.

The contributions of this study are summarized as follows:
- Proposing SGLC for Dehazing High-Resolution images without compromising global or local features. SGLC has two successive blocks: the Global Features Generator (GFG) block and the Local Features Enhancer (LFE) block.
- Proposing the Grid Patching process to capture the global features from High-Resolution images.
- Customizing a loss function to enable both the Dehazing and the Enhancer models to learn better the High Frequencies components.
- Proving that putting GFG before LFE works better than the reverse process. This follows the intuitive rule: generate the global content, then enhance the small fine details.

## 2. Related works

Handling the Dehazing operation for High-Resolution Images was the aim of many works in literature. Among the first works on this scope, Sim et al.[27] designed the Dehazing Generative Adversarial Network (DHGAN). They trained the generator on hazy patches of input images scaled to reduced sizes. This helped the model to capture more global features. They also modified the cross-entropy loss to include multiple outputs. In another work, Ki et al. [28] proposed BEGAN (Boundary Equilibrium Generative Adversarial Network). They enlarged the receptive field and trained the discriminator on High-Resolution images. The images were conditioned on downscaled hazy images. Moreover, Bianco et al. [29] designed HR-Dehazer (High-Resolution Dehazer). The HR-Dehazer is based on an Encoder-Decoder architecture and a specifically designed loss. This enables the network to learn the semantics of the clean image and to improve the consistency in local structures. They made the architectures scale invariant and able to work on large sizes. Besides, Zheng et al. [19] treated the same problem differently. They conceived a new model, the 4k-Dehazer, incorporating three networks working jointly on a bilateral space. Every one of them feeds it and gets features from it simultaneously. The global architecture is efficient despite the High-Resolution of the image. They constructed a large-scale 4K image dataset to assess the model and found that it performs favorably compared to the state-of-the-art Dehazing models. In another tentative, Chen et al. [30] designed the H2RL-Net network composed of two branches. The first focuses on the High-Resolution details, with the second collects the semantic features using a complementary set of multiresolution CNN streams. They exploited the PCF (parallel cross-scale fusion) module. Which incrementally aggregates features from many scales at the specific resolution level and exchanges information from top to bottom and from bottom to top. Also, they used the CFR (Channel Feature Refinement) block, which recalibrates the channel-wise features.

From the above, we note that all the works that targeted High-Resolution Image Dehazing focused on designing new architectures that meet the peculiarities of the large images. Although efficient, these works push towards using specific architectures for High-Resolution cases instead of the well-developed and tested architectures designed for typical Dehazing cases. Therefore, we need a generic framework that allows using them by extending their efficiency to large images. This study targets this problem in the Dehazing state of the art and aims to bridge the gap between the Dehazing architectures and the High-Resolution images. Moreover, when trying to combine global and local features, all the methods stated above conceived a parallel way for it, either by designing a bilateral latent space or by working on multiple networks run in parallel on the input image. Our work solved the problem differently by conceiving a streamlined process working sequentially on the global features (using the GFG block) and then working on the local features (using the LFE block). The blocks are discernable, which helps to estimate better the performance of every block apart. Also, it helps to work on them in a modular way and to improve them in the following research better works.

## 3. Proposed Methodology

This section introduces the different parts of the SGLC framework: Streamlined Global and Local Features combinator. A global diagram representing the framework is displayed in Fig.1. The diagram depicts the two blocks constructing SGLC, the Global Features Generator (GFG) and the Local Features Enhancer (LFE). In the following subsections, the entire process is explained.

### 3.1. Global Features Generator (GFG)

First, the image entered into SGLC passes to the first block: The Global Features Generator (GFG). This block aims to generate the global content accurately by considering the whole image content. However, because of the High-Resolution, we need a new method to collect them in a systematic way to feed the network without extra computational cost. Thus, we introduce in this study the Grid patching method.

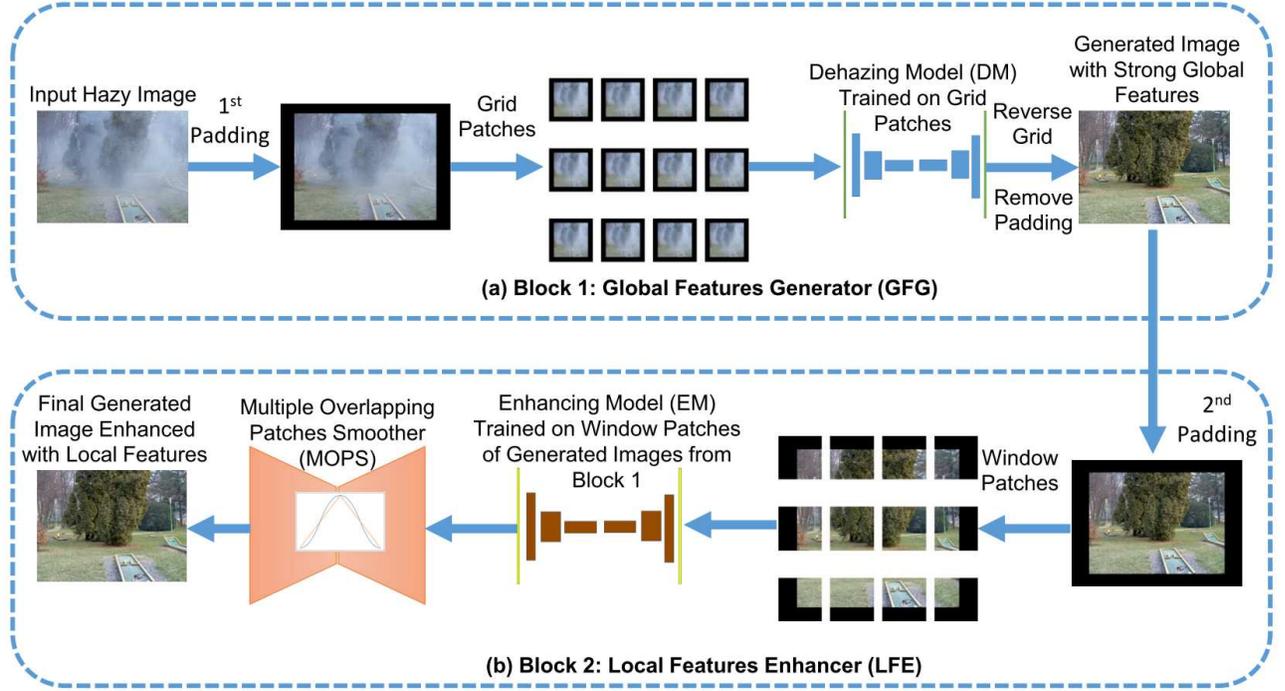

Figure 1: SGLC diagram

*3.1.1  Grid patching for global features learning*

Let us consider first $G \times G$ the size of the patches we want to generate. We are considering square sizes because this is the case in most networks, but the method can also be adapted to rectangular sizes.

If we consider an input image $I$ with height $H$ and width $W$, the Grid patching begins by adding extra padding so that the $H$ and $W$ will be dividable by $G$. Let us consider $H'$ and $W'$ the size of the padded image $I'$. If $H$ is already dividable by $G$, then $H' = H$. Also, if $W$ is dividable by $G$, then $W' = W$. Otherwise, $H$ and $W$ are calculated using the following equations:

$$H' = ((H \ div \ G) + 1) * G \quad (3)$$

$$W' = ((W \ div \ G) + 1) * G \quad (4)$$

Let us consider $n_h = H'/G$ the height number of divisions, and $n_w = W'/G$ the width number of divisions. The total number of patches generated at the end will be:

$$N = n_h * n_w \quad (5)$$

The Grid patching method works on the padded $I'$ so that for every selected pixel, we jump $n_h$ vertically to select the next vertical pixel. If we want to select the next horizontal pixel, we also jump horizontally $n_w$ pixels. In the end, every patch takes equally distanced pixels from the whole image. Equal distances are considered from both the width and the height. The complete process can be explained in Algorithm 1, where from the padded input image $I'$ we generate $N$ Grid patches. For every patch $P_k$, we first initialize it into a zeros-valued three-channel matrix. Then, we select the pixels from $I'$ by keeping the equal vertical and horizontal distances described above.

| Algorithm 1: Grid Patching |
|---|
| **1: Input** |
|     1. $I'$: The padded image |
|     2. $n_h$: The height number of divisions |
|     3. $n_w$: The width number of divisions |
|     4. $N$: The number of generated patches |
| **1: Grid patching** |
|   i.   For $P_k$, $k$ in (0… $N-1$): // P: a generated patch |
|     a.   $P_k \leftarrow$ **zeros** (G, G,3) // initialization to zeros |
|     b.   For $i$ in (0 … $G-1$): |
|       i.   For $j$ in (0 … $G-1$): |
|         1.  $P_k[i,j] \leftarrow I'[i + n_w, j + n_h]$ |
|         2.  End |
|       ii.  End |
|     c.   End |

The Grid patching algorithm is applied for every image fed to the Dehazer Model (DM) either in the training or the inference. More details will be explained in the next paragraph.

### 3.1.2 Dehazing Model (DM)

The Dehazing Model works to understand global features. Therefore, it is only trained on Grid Patches. Every dataset used is preprocessed using the Grid Patching method so that it will learn how to grasp the global features, whatever the input size.

In SGLC, we used Uformer [26] in both the Dehazing and Enhancer Models. Uformer is a U-shaped hierarchical network that employs skip connections between the encoder and decoder, like the U-Net model. Given a hazy input image $i_h \in \mathbb{R}^{c \times h \times w}$, the network initially employs a $3 \times 3$ convolutional layer with a LeakyReLU activation function to extract low-level features $x_i$. The resulting feature maps $x_i$ are then passed through the encoder stages, where each stage comprises a stack of LeWin Transformer blocks that leverage a locally enhanced window mechanism to capture long-range dependencies while minimizing computational cost. This mechanism is achieved by utilizing non-overlapping windows during self-attention computations on the feature maps. Moreover, the encoder incorporates a down-sampling layer that reduces the spatial dimensions of the feature maps while simultaneously increasing their number of channels. The down-sampling layer reshapes the flattened features into 2D spatial feature maps and subsequently performs down-sampling by doubling the number of channels through a $4 \times 4$ convolutional layer with stride 2.

The LeWin block utilizes the self-attention mechanism to capture long-range dependencies in the feature maps. Additionally, it employs the convolution operator in the Transformer to capture the necessary local context. This approach enables the block to capture both global and local information effectively, making it a suitable choice for various image restoration tasks. The LeWin Transformer block is composed of several key components, including Non-overlapping Window-based Multi-head Self Attention (NW-MSA), Locally Enhanced Feed-Forward Network (LFF), and Layer Normalization (LN). LeWin can be represented mathematically as:

$$X'_l = NW\text{-}MSA(LN(X_{l-1})) + X_{l-1} \quad (6)$$

$$X_l = LFF(LN(X'_l)) + X'_l \quad (7)$$

where $X'_l$ is the output of the $NW\text{-}MSA$ module and $X_l$ is the output of $LFF$ module. The self-attention mechanism used in NW-MSA differs from the global self-attention used of the vanilla transformer in that self-attention is implemented within non-overlapping local windows, which reduces the computational cost. Given features map $x_i$ with height and width $(h \times w)$, then the $x_i$ is divided into non-overlapping windows of $m \times m$ size, and the number of windows is calculated by $\frac{h*w}{m^2}$. The computation of multi-head self-attention in non-overlapping windows can be expressed as follows:

$$x = \{x_1, x_2, x_3, \dots, x_n\}, \; n = \frac{h*w}{m^2} \quad (8)$$

$$y_k^i = Attention(x_i w_k^Q, x_i w_k^K, x_i w_k^V), i = 1,2,\dots,n \quad (9)$$

where $x_1, x_2, x_3, \dots, x_n$ are non-overlapping windows, $w_k^Q, w_k^K, w_k^V$ are the projection matrices of the queries, keys, and values, $k$ is the number of heads, $y_k^i$ is the output of the attention process. Relative position coding bias $B$ was applied in the attention module inspired by[31], [32]. The process of attention can be expressed as follows:

$$Att\,(Q, K, V) = softmax\left(\frac{QK^T}{\sqrt{d_k}} + B\right)V \quad (10)$$

In the LFF, a linear projection layer is applied to each token to increase its feature dimension. The tokens are then resampled into 2D feature maps, and a 3×3 depth-wise convolution is employed to capture local information. Subsequently, the features are flattened back into tokens, and the channel dimension is reduced using another linear layer to match the input channel dimension. This process enhances the local context information by applying a convolution operator in the transformer model.

The decoder of Uformer is composed of multiple stages, each of which incorporates an up-sampling layer and several LeWin Transformer blocks, similar to the encoder. Specifically, the output of each encoder stage is connected sequentially with the highest sample features from the previous decoder stage via a skip connection. This enables the decoder to access and combine features from different scales, thereby facilitating the generation of high-quality restored images.

The Uformer represents one state-of-the-art architecture in the Image Restoration domain. We note the particular concern given to the combination of local and global features in the design. Although compromising the Multi-Head Self Attention (MSA) by only restricting it to Non-Overlapping Windows, the effect was minimal on the efficiency, and it reduced the computational cost. However, scaling this architecture to High-Resolution makes it lose its inherent features in handling global and local features. SGLC helps to scale any successful Image Restoration model to work better for High-Resolution images.

### 3.1.3 Self-Supervised Learning

To enhance the learning capability of both the Dehazing Model (DM) and the Enhancer Model (EM), we used self-supervised learning as a first step before beginning the training process. Similar to what has been done in [33], [34], we constructed from the clean images a new dataset. In this dataset, we create from every image a new image where we randomly draw small in-painted squares. The model will be trained on how to reconstruct the original image from these corrupted versions, which helps it to improve its learning of the underlying representation of the data. This method was previously used in literature in Context Encoders [34] and Contextual Attention [33]. The idea was to train the model to fill data based on the global image context. They found that it enhances the model's capability to handle supervised Image Restoration tasks later. However, in the Dehazing task, the in-painted squares should be white to imitate haze patterns better. Geometric forms other than the square form could also be added.

### 3.1.4 Customized Loss function

To train the DM and the EM models, we used the following customized loss function:

$$\mathcal{L} = \sqrt{\|I - \hat{I}\|^2 + \|\Pi(I) - \Pi(\hat{I})\|^2 + \varepsilon^2} \quad (11)$$

Where $I$ is the ground truth image, $\hat{I}$ represents the predicted image, $\Pi$ represents the Laplacian Pyramid operator, $\varepsilon$ is a constant empirically set to $10^{-3}$ for all the experiments.

The loss function contains three components. The first is the comparison of the original image with the predicted one $\|I - \hat{I}\|^2$. This component helps to learn the spatial content of the image. The second component works on another image representation: the Laplacian Pyramid, which was introduced by[35]. This representation has been used a lot for image encoding and compression. It focuses on the edge information in the data and emphasizes the structural content. The second component compares the Laplacian Pyramid of the original images with the predicted one: $\|\Pi(I) - \Pi(\hat{I})\|^2$. This term helps to mitigate inherent spectral bias [36] in Deep Neural Networks and to better learn the High-Frequency components, which are very important in the Image Restoration domain.

The third component is the $\varepsilon^2$, and it is inspired by the Charbonnier penalty function [37]:

$$\rho(x) = \sqrt{x^2 + \varepsilon^2} \quad (12)$$

This penalty function helps to improve the robustness of the $L1$ loss by squaring the value and adding a very small constant ε, and at the end, putting all of them under the square root. This makes the Image Restoration more robust[38] by preventing the loss from vanishing when its value approaches 0.

### 3.1.5 Reverse Grid Patches Reconstruction

During the inference, the input image passes first by the Grid Patching preprocessing step. Then, the patches are passed separately to the network, either in batches or one by one. The predicted set of Patches $\widehat{P}_l, l \in 0..(N-1)$ are used to reconstruct the full predicted image $\widehat{I}_F$ by using the following Algorithm:

| Algorithm 2: Reverse Grid Patches Reconstruction |
|---|
| **1: Input** |
|    1. $\widehat{P}_l, l \in 0..(N-1)$: The predicted Grid patches |
|    2. $\widehat{I}_F$: The full predicted image to be generated |
|    3. $n_h$: The height number of divisions |
|    4. $n_w$: The width number of divisions |
|    5. $N$: The number of generated patches |
| **1: Reconstruction** |
|   i.   $\widehat{I}_F \leftarrow zeros\,(G * n_w, G * n_h, 3)$ |
|   ii.   For $\widehat{P}_l, l \in 0..(N-1)$: |
|     a.   $i_o \leftarrow l\ mod\ n_w$ //horizontal offset |
|     b.   $j_o \leftarrow l\ div\ n_w$ // vertical offset |
|     c.   For $i_p$ in $(0 \dots G-1)$: |
|       i.   For $j_p$ in $(0 \dots G-1)$: |
|         1.   $x \leftarrow i_o + i_p * n_w$ |
|         2.   $y \leftarrow j_o + j_p * n_h$ |
|         3.   $\widehat{I}_F[x,y] \leftarrow \widehat{P}_l[i_p, j_p]$ |
|         4.   End |
|     ii.   End |
|     d.   End |

The image generated by Algorithm 2 will be unpadded to conform to the original image size and to serve as the first dehazed image generated by the first block GFG. Then, it will be entered as input to the second block LFE.

## 3.2. Local Features Enhancer (LFE)

The LFE block serves to enhance the image quality regarding the local features. The image generated by GFG will be divided into Window patches and passed to the Enhancer Model (EM).

### 3.2.1 Enhancer Model (EM)

Compared to the DM model, the EM model has two main characteristics. The first is that it works on Window patches, not Grid patches. The second characteristic is that it is trained on the predicted dehazed images from GFG with their clean counterparts. In more detail, all the images in the dataset must be first predicted by the first block. Then, they are padded similarly to what has been done in the first block so that the new size will be dividable by the Window size. Then, these images will be divided into Window patches. These patches, alongside the Window patches of the clean counterparts, will form a new dataset for training EM.

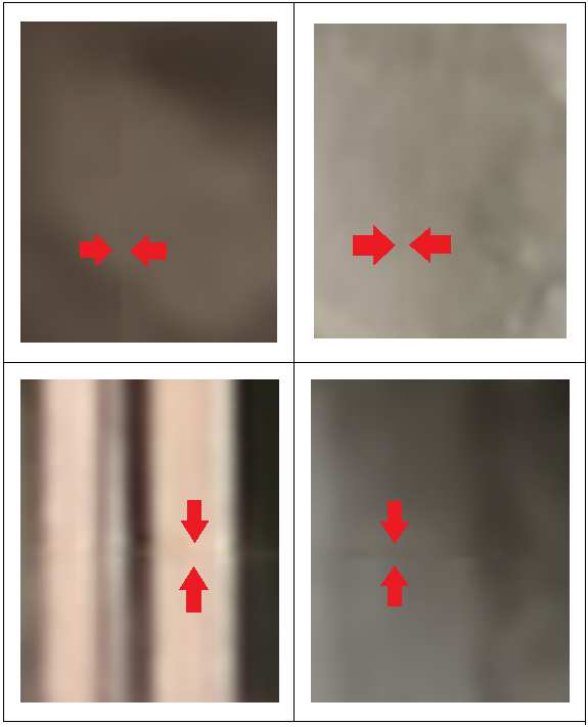

Figure 2: The Vignetting artifacts when using EM with the Reverse Window Patches Reconstruction algorithm

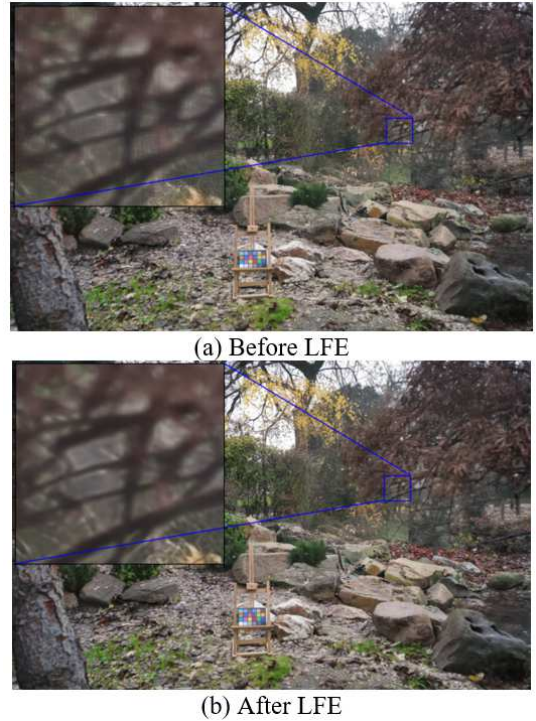

Figure 3: Predicted image before and after LFE block

The same Uformer model described above was used in EM. Also, the same self-supervised strategy, as well as the same customized loss function, were used in EM. The goal of the EM is to learn how to deal with the shortcomings in the first dehazed image using local features grasped from spatially continuous data patches (the Window patches).

In the inference stage, we will have some visual artifacts if we use a standard Reverse Window Patches Reconstruction algorithm to generate the full-sized predicted image from the predicted patches. Hence, we used a more advanced algorithm described in the next section.

### 3.2.2 Multiple Overlapping Patches Smoother (MOPS)

The Reverse Window Patches Reconstruction algorithm generates images that suffer from Vignetting artifacts visible as a discontinuity in the spatial domain between adjacent patches. Four examples of these vignetting artifacts are displayed in Fig.2. Although these artifacts are sometimes hard to detect, they are reducing the total Dehazing efficiency, as demonstrated in the Experimental part. These types of artifacts are due primarily to an underlying limitation in the CNN networks themselves [39], such as the difficulty in ensuring translational equivarance because of using the strides and the zero-padding operations.

Thus, to solve this issue, the LFE applies a blending algorithm [40]. This algorithm aggregates multiple overlapping predictions of the crops by the intermediate of a window function[41]. This algorithm was designed first to improve the quality of Semantic Segmentation maps [40], [41]. In this study, we applied it in the Dehazing context and identified it for simplicity as Multiple Overlapping Patches Smoother (MOPS).

Initially, MOPS applies rotations and mirroring on the original full-sized input image to better ensure the invariance of the prediction over them. Then, for every one of these full-sized images, it sums multiple overlapping adjacent predictions with an overlap degree of 50% using a simple second-order spline window function.

The main drawback of MOPS is its computational cost. Therefore, we did in the Experimental part an ablation study to estimate its utility compared to its cost.

Finally, the image generated by MOPS and EM is the final dehazed image supposed to ensure better local details enhancement. To demonstrate this improvement, Fig.3 shows a predicted image before and after applying the LFE block. As shown in the image, LFE enhanced the pixel and tiny details. In addition, some minimal artifacts are removed, and the quality approaches the quality of the ground truth. The improvements are emphasized in the metrics described in the Experimental part.

## 3.3. SGLC algorithm

Finally, the inference stage of SGLC is detailed in the following Algorithm:

| Algorithm 3: SGLC inference stage |
| --- |
| **1: Input** |
|    1. $I$: The hazy input image |
|    2. $N$: The number of generated patches |
| **1: Block 1 (GFG)** |
|   i. $I' \leftarrow padding(I)$ // adding padding |
|   ii. $\{P_k, k \text{ in } (0..N-1)\} \leftarrow Grid\_Patching(I')$ |
|   iii. $\{\widehat{P_l}, l \in 0..(N-1)\} \leftarrow DM(\{P_k, k \text{ in } (0..N-1)\})$ |
|   iv. $\widehat{I_F} \leftarrow RGPR(\{\widehat{P_l}, l \in 0..(N-1)\})$ // Reverse Grid Patching Reconstruction |
|   v. $\widehat{I'_F} \leftarrow unpadding(\widehat{I_F})$ |
| **1: Block 2 (LFE)** |
|   i. $IM = padding(\widehat{I'_F})$ // second padding |
|   ii. $\{W_i, i \text{ in } 0..(N-1)\} \leftarrow Window\_Patching(IM)$ |
|   iii. $\widehat{IM} = MOPS(EM, \{W_i, i \text{ in } 0..(N-1)\})$ |
|   iv. $\widehat{IM'} = unpadding(\widehat{IM})$ //Final predicted image |

## 4. Experiments

In this section, the performance of SGLC is compared against the original performance of Uformer applied on a resized image without SGLC. Also, we compare it against DW-GAN[6], which is the winner of the NTIRE Non-Homogeneous Dehazing Challenge of 2021[5]. We also compared it with 4k-Dehazer [19], a state-of-the-art Dehazing algorithm explicitly designed for High Resolution. Images for 4K-Dehazer were resized to 4000*4000. We could not work on original sizes due to memory limitations. The scale of the images used in the challenge makes many Dehazing models for High-Resolution images practically useless due to GPU memory limitations. In contrast, the SGLC does not care how large the image is. The exact process works for any image size.

### 4.1. Experimental configurations

*4.1.1  Dataset description*

The dataset used for assessing the validity of the SGLC is released within the HR Non-Homogeneous Dehazing Challenge of the NTIRE CVPR 2023 Workshop. They named it HD-NH-HAZE. The dataset contains 50 images with huge sizes (either 4000*6000 or 6000*4000). Of them, 40 are reserved for training, 5 for validation, and 5 for the test. The challenge organizers keep the clean image of the validation and test datasets private. Thus, we divided the original 40 images into 36 for training our model and 4 for testing it. No extra data were used in our study.

*4.1.2  Implementation details*

For running SGLC, we used a Lambda AI server having: 8 GPUs NVIDIA QUADRO 8000. Every GPU has 48 GB of GDDR6 video memory. The server has 512 GB of RAM and two processors Intel Xeon Silver 4216, every one of them has 16 cores. Another workstation having one GPU, NVIDIA QUADRO 8000, is also used.

### 4.2. Results and Analysis

The results of applying different models in our local training are displayed in Table 1. All the results consider a patch size of 1024*1024 for both the Grid Patching and the Window patching processes.

Table 1: Performance of Uformer, DW-GAN, 4k-Dehazer, SGLC, and Inv-SGLC on the HD-NH-HAZE (our test set)

| Model description | PSNR (dB) | SSIM | Inference time (sec/image) |
| --- | --- | --- | --- |
| Uformer applied on the resized image of 1024*1024 | 14.90 | 0.6403 | 14.9 |
| 4K-Dehazer | 17.46 | 0.7230 | 9.0 |
| DW-GAN | 17.48 | 0.7367 | 45.7 |
| SGLC-GFG only | 24.49 | 0.8176 | 46.6 |
| SGLC without MOPS | 25.38 | 0.8511 | 86.0 |
| SGLC (GFG +LFE) | **25.43** | **0.8524** | **553.5** |
| Inv-SGLC LFE only without MOPS | 23.07 | 0.8300 | 40.0 |
| Inv-SGLC LFE only | 23.25 | 0.8335 | 556.1 |
| Inv-SGLC (LFE + GFG) | 24.39 | 0.8392 | 605.7 |

As shown in the table, the original Uformer applied on the resized version of all the images to 1024*1024 got weak performance. Also, DW-GAN and 4k-Dehazer got weak results. These results demonstrate that the Dehazing model relies on fine and pixel details and works better with the original format of images without scaling or resizing.

Regarding the SGLC pipeline, the big jump in the results is made by the GFG block. Then, applying the LFE without MOPS increases the PSNR by a margin of 0.89. At the same time, the SSIM was increased by a margin of 0.0335. Therefore, the margin is significant, especially for the SSIM. For the application of MOPS, the improvement was 0.05 in the PSNR and 0.0013 in the SSIM. Therefore, the improvement made by MOPS is not significant. However, it can be considered for exceptional cases where we need the best accuracy, no matter the computation cost.

For the computational cost, the whole pipeline of SGLC needed 553.5 sec for one image of the size of 4000*6000.

The time is 46.6 sec for the GFG block, 39.4 sec for the LFE block, and 467.5 sec for MOPS. Therefore, in most cases, the application of SGLC without MOPS will be sufficient as it holds the most performance of the SGLC.

To assess if changing the order of GFG and LFE matters, we tried the inverse process of SGLC (Inv-SGLC in Tab.1). This pipeline begins by working on local features by training the EM model on window patches of hazy/clean images. Then MOPS is applied to smooth the EM predictions. After that, the image passes to the Grid patching process to be fed to the DM model trained on Grid patches of LFE_predicted/clean patches. The result shows that the global performance of Inv-SGLC is remarkably below SGLC.

Also, we note that the improvement made by LFE has more impact on SSIM than PSNR. In addition, Fig.4 helps to assess visually the image quality generated by SGLC against the image quality generated by DW-GAN and the image generated by the Uformer model displayed on the first row of Tab.1. We can see the superiority of the SGLC compared to them.

Table. 2 displays the final leaderboard results in the NTIRE 2023 Non-Homogeneous Dehazing Challenge report [42]. SGLC was ranked 5th among 17 submitted solutions based on four metrics: PSNR, SSIM, LPIPS, and MOS.

Table 2: Final leaderboard results in the NTIRE 2023 Non-Homogeneous Dehazing Challenge [42]

|  | PSNR | SSIM | LPIPS | MOS |
| --- | --- | --- | --- | --- |
| SGLC performance | 22.27 | 0.7 | 0.439 | 7.4 |
| Best performance among No Extra data solutions | 22.27 | 0.7 | 0.384 | 7.65 |
| SGLC Rank among No extra data solutions | 1/12 | 1/12 | 7/12 | 2/12 |
| Best performance among all solutions | 22.96 | 0.71 | 0.345 | 8.07 |
| SGLC Rank among all solutions | 3/17 | 4/17 | 10/17 | 5/17 |

SGLC's maximum performance is among the solutions that did not use extra data. It is the best in PSNR and SSIM and the second in MOS. Regarding all the solutions, SGLC was ranked 3rd in PSNR, with a gap of 0.69 compared to the best. Also, it was ranked 4th in SSIM, with a gap of 0.01 compared to the first. The main limitation of SGLC is recorded in the LPIPS metric, which should be considered during the following improvements. Also, using extra data can improve the results.

## 5. Conclusion

In this study, the SGLC framework is proposed. This framework helps any Dehazing model to scale its performance to High-Resolution images. It begins by generating the first dehazed image with robust global features using the GFG block. This block learns global feature information by training the Dehazing Model (DM) on Grid Patches of the hazy images. Then, the generated image will pass through the LFE block to enhance the image by the intermediate of the local features. A customized loss alongside the Self-Supervised Learning and the MOPS were used jointly to improve the global performance of SGLC. Also, SGLC is better than Inv-SGLC. This superiority demonstrates that working on the global features at the first stage is better. Regarding the limitations, SGLC suffers from high computational costs, especially when using MOPS. Also, the streamlined process, which has many advantages, currently hinders its parallelization and adoption for video processing. Also, we need to work on other enhancements, such as hand-crafted priors, or investigate the use of Diffusion Models, which recently attracted researchers' attention in the Image Synthesis domain.

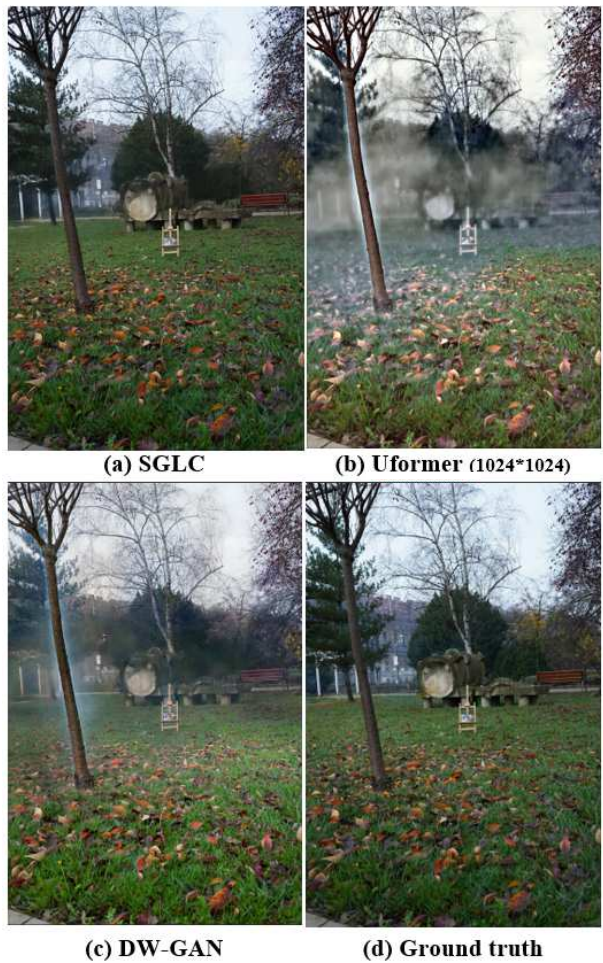

Figure 4: Dehazing performance of SGLC compared to Uformer (1024*1024), DW-GAN, and the Ground Truth.

**Acknowledgment:** The authors thank Prince Sultan University for their support and funding in conducting this research.